# On defining 'I'

## "I -logy"

By: Farzad Didehvar

Email: didehvar@aut.ac.ir

**Abstract:** Could we define "I"? Throughout this article we give a negative answer to this question.

More exactly, we show that there is no definition for "I" in a certain way. But this negative answer depends on our definition of definability. Here, we try to consider sufficient generalized definition of definability. In the middle of paper a paradox will arise which makes us to modify the way we use the concept of property and definability.

Here, definability is defined based on the properties of objects and possessing the properties by objects.

Throughout this article we don't define and consider property as precise as a mathematical

Definition (by property here, we mean the property of a cognitive agent), but we try to make it as precise as possible.

Indeed we suppose that, our intuition about property is sufficient to develop the subject here.

**Definition**: p-sentences are the set of sentences which are constructed by induction as following:

1. Every sentence in the form of "X has property p" is a p-sentence

2. The logical combinations of p sentences are p-sentence.

Here, a **definable object** is an object which there is a p-sentence which defines and specifies (****) that object.

**Definition:** For every property p, by A(P) we mean the set of objects which has the property p.

The following definitions and theorems are based on the definitions and theorems in [1]. Possibly,

In the first glance the subject seems a little odd, unless we read [1] first.

**Property *:** A property p which among the objects of A (P) (either in a real world or in a possible world) there is an element (a cognitive agent one) which denies " I "belongs to A (p) (by simply saying a sentence, true or not true", so either he denies any agreement to "I" or he is not able to have such an agreement (stubborn agent)), is a property'. (So he could point "I" and call it somehow).

Moreover, if a property' is as powerful as any cognitive agent which belongs to A (P') is comparable to "I", we call it property*.(Two cognitive agents are comparable if and only if their claims are comparable. About the role of comparability, in [٢] it is shown that comparability between cognitive agents is a central concept). (****)

**Example:** Any possible definition of "I" is a property*(more explanation in theorem٢).

**Example**: To be a machine and "to be a human being" "not to be machine" and "not to be human being" are property*.

Since, there is a machine which one of its outputs is: "I" is not a machine.

Also, in a possible world there is a human being who says that: "I" (me) is not a human being.

**Definition:** Any property about cognition of agents is called here cognition property.

**Thesis**: Any sufficient powerful cognition property q is a property'. (i.e. Cognition property is stronger than Property').

It means that, any enough complicated cognitive agents who possesses the property q, could deny "I" belongs to A(q).

Now, we are going to prove that:

"I" is something special.

In other words, there is no property* for I, in a certain way.

To do this we prove the following theorems, but before starting that we remind here, the proof is similar to the proofs in [١], [٢]. In fact, in the references [١] and [٢], the material provides a proof stronger than what we claimed there. The stronger statement is:

If we demonstrate that "I am a machine", we will be as much certain as I am a machine that we will be certain I am not a machine.

**Theorem ١:** We can't be certain that "I" has some property* like q, moreover by considering that "I'' has

Property* like q, we are as certain that it has this property* as it hasn't this property.

**Proof:** (By following the idea in reference [١]). In brief, if "I "has some property*, by definition there is another cognitive agent among the elements which has this property which denies that I has property* (Stubborn agent, stubborn agent could be merely in a possible world and not in a real world). The main question here is: Who is right? I, who belives possessing the property* or stubborn agent. (by definition their claims are comparable.) Any judgment needs a faire place. But any faire place and faire rules are decided by "I" and stubborn agent either denies that or he has no idea about. So, no judgment would be faire, it concludes that "I" couldn't judge who is wrong, "I" himself or stubborn agent. Therefore "I" can't be certain that he has this property*. Moreover he is certain that "I" is true as much as that stubborn agent is true. So, we have the conclusion.

Note that, when we are sure about sentence r, implicitly we deny people who deny r.

**Theorem ۲:** Any property (p-sentence) which can define "I" and also "I" can phrase it, is a property*.

**Proof:** Since "I" could say "I have not this property", so by definition of property* we have the above theorem (there is a stubborn agent). (This ability of "I" is considered here as minimum cognition power of "I").

**Conclusion:** No property (p-sentence) can define "I", in a certain way. Moreover we are certain that p is the property of I as much as it is not so.

**Proof:** Since any such property should be property* (by Theorem ۲), but by theorem ۱ we can't be certain that "I" has this property, so no property could describe "I" in a certain way.

So, not only "I" is not definable, but also it is not even describable, it is something quite special.

**Remark**: "I" has no enough powerful cognition property, in a certain way. (A property*)

**Paradox:** Consider the identity property id. A(id) are the things which are themselves. "I is I", but some agent T could say I is not I, and T believes that.

So, I can't be sure that he is himself (even with the similar scenario "I" am "I"), which is a quite contra intuitive assertion. Is it true to say that our theory is collapsed, somehow?

Above paradox asserts that our modeling doesn't adapt completely to the reality, there is a point here which is neglected.

In following, we try to resolve the difficulties.

**Refuting the paradox:** *(Inner sight, outer sight)* In fact we have two types of watching the world, outer sight: the way which we know the world as an observer, inner sight: The way in which we feel (ourselves or something), like the way which we feel pain or we feel ourselves. (The inner sight provides the mediate knowledge and the outer sight provides the intermediate knowledge).              .

Now this distinction makes the following point, which focus on comparability.

Knowing ourselves and feeling that "I am I" is from inner sight and when someone says something about me it is from outer sight and these two points of view are not comparable, Even though he speaks about my inner sight problem.( in the condition, it is supposed he has right to do that).

So in theorem ١ we are not able to compare the assertion of "I" is "I" by stubborn agent and "I" am "I" by I. As a conclusion we could have no judgment between these two assertions.

So, the main principal we use here, in order to resolve complexities is, "Do not compare the assertions from inner sight and outer sight in any judgment, they are not comparable ".

This point do not harm our previous assertion in [١], since all assertions there, they were from outer sight.

Here, if we assert ' "I" is "I" ' by inner sight, it is not comparable by stubborn agent's assertion. But if we assert it by outer insight they are comparable, so to assert this sentence in this view could makes (****) some ambiguities.

We abbreviate it as follows:

1. ١. We are able to compare two outer sight assertions,
2. ٢. We are able to compare two inner sight assertion, (Although I am quite doubtful about this assertion),
3. ٣. Here we accept that an inner sight assertion and an outer sight assertion are not comparable.

There are some evidences which support the third claim, besides the above discussions. Usually, we believe to the certainty of inner sight. It seems, if there is any certainty, it is in inner sight. But we haven't the same idea for outer sight. Doubt and outer sight have a long live brother hood.

In chapter ٢, by considering the above assertion we will rephrase the arguments again.

Note that again comparability between assertions comes in to play in theorem ١, in judgment between assertions.

**CHAPTER ٢**

**Remark:** In this chapter we restrict ourselves to outer sight assertions. So, any sentence and property here is an outer sight property (outer property). Now we repeat what we have written word by word, in addition to the mentioned considerations, the exception is two last phrases, so you can jump to the last phrases immediately.

Here, definability is defined based on the properties of objects and possessing the properties by objects.

Throughout this article we don't define and consider property as precise as a mathematical

Definition (by property here, we mean the property of a cognitive agent), but we try to make it as precise as possible.

Indeed we suppose that, our intuition about property is sufficient to develop the subject here.

**Definition**: p-sentences are the set of sentences which are constructed by induction as following:

1. Every sentence in the form of "X has property p" is a p-sentence

2. The logical combinations of p sentences are p-sentence.

Here, a **definable object** is an object which there is a p-sentence which defines and specifies (****) that object.

**Definition:** For every property p, by A(P) we mean the set of objects which has the property p.

The following definitions and theorems are based on the definitions and theorems in [1]. Possibly,

In the first glance the subject seems a little odd, unless we read [1] first.

**Property *:** A property p which among the objects of A (P) (either in a real world or in a possible world) there is an element (a cognitive agent one) which denies " I "belongs to A (p) (by simply saying a sentence, true or not true", so either he denies any agreement to "I" or he is not able to have such an agreement (stubborn agent)), is a property'. (So he could point "I" and call it somehow).

Moreover, if a property' is as powerful as any cognitive agent which belongs to A (P') is comparable to "I", we call it property*.(Two cognitive agents are comparable if and only if their claims are comparable. About the role of comparability, in [2] it is shown that comparability between cognitive agents is a central concept). (****)

**Example:** Any possible definition of "I" is a property*(more explanation in theorem2).

**Example**: To be a machine and "to be a human being" "not to be machine" and "not to be human being" are property*.

Since, there is a machine which one of its outputs is: "I" is not a machine.

Also, in a possible world there is a human being who says that: "I" (me) is not a human being.

**Definition:** Any property about cognition of agents is called here cognition property.

**Thesis**: Any sufficient powerful cognition property q is a property'. (i.e. Cognition property is stronger than Property').

It means that, any enough complicated cognitive agents who possesses the property q, could deny "I" belongs to A(q).

Now, we are going to prove that:

"I" is something special.

In other words, there is no property* for I, in a certain way.

To do this we prove the following theorems, but before starting that we remind here, the proof is similar to the proofs in [1], [2]. In fact, in the references [1] and [2], the material provides a proof stronger than what we claimed there. The stronger statement is:

If we demonstrate that "I am a machine", we will be as much certain as I am a machine that we will be certain I am not a machine.

**Theorem 1:** We can't be certain that "I" has some property* like q, moreover by considering that "I'' has

Property* like q, we are as certain that it has this property* as it hasn't this property.

**Proof:** (By following the idea in reference [1]). In brief, if "I "has some property*, by definition there is another cognitive agent among the elements which has this property which denies that I has property* (Stubborn agent, stubborn agent could be merely in a possible world and not in a real world). The main question here is: Who is right? I, who belives possessing the property* or stubborn agent. (by definition their claims are comparable.) Any judgment needs a faire place. But any faire place and faire rules are decided by "I" and stubborn agent either denies that or he has no idea about. So, no judgment would be faire, it concludes that "I" couldn't judge who is wrong, "I" himself or stubborn agent. Therefore "I" can't be certain that he has this property*. Moreover he is certain that "I" is true as much as that stubborn agent is true. So, we have the conclusion.

Note that, when we are sure about sentence r, implicitly we deny people who deny r.

**Theorem 2:** Any property (p-sentence) which can define "I" and also "I" can phrase it, is a property*.

**Proof:** Since "I" could say "I have not this property", so by definition of property* we have the above theorem (there is a stubborn agent). (This ability of "I" is considered here as minimum cognition power of "I").

**Conclusion:** No property (p-sentence) can define "I", in a certain way. Moreover we are certain that p is the property of I as much as it is not so.

**Proof:** Since any such property should be property* (by Theorem 2), but by theorem 1 we can't be certain that "I" has this property, so no property could describe "I" in a certain way.

So, not only "I" is not definable, but also it is not even describable, it is something quite special.

**Remark:** "I" has no enough powerful cognition property, in a certain way. (A property*)

**Question:** Could we speak about a science in the name of "I-logy"?

***Acknowledgment:*** *The writer would like to thank Dr Abbasian and Dr Vahid for some useful discussions which we had.*